\documentclass{article}
\usepackage{spconf,amsmath,graphicx, xcolor, enumitem, pifont, makecell, tabularx, multirow, booktabs, colortbl, amssymb, tikz, arydshln, xcolor,float, caption,indentfirst, nccmath}
\usepackage[colorlinks=true, urlcolor=black]{hyperref}
\usepackage[skip=4pt]{caption}
\usepackage{stfloats}
\newcommand{\cmark}{\ding{51}}

\title{Enhanced Detection of Tiny Objects in Aerial Images}
\name{
\begin{tabular}{ccc}
Kihyun Kim\sthanks{Thanks to Hyesoon Kim and Seonhwan Kim for funding \& support}$^{1}$ & Michalis Lazarou$^{2}$ & Tania Stathaki$^{1}$ \\
yoo07220@gmail.com & michalislazarou93@gmail.com & t.stathaki@imperial.ac.uk
\end{tabular}
}
  
\address{$^{1}$ Dept. of Electrical \& Electronic Engineering, Imperial College London \\
      $^{2}$ Center for Vision, Speech and Signal Processing, University of Surrey}
\begin{document}
%
\maketitle
\begin{abstract}
While one-stage detectors like YOLOv8 offer fast training speed, they often under-perform on detecting small objects as a trade-off. This becomes even more critical when detecting tiny objects in aerial imagery due to low-resolution targets and cluttered backgrounds. 
To address this, we introduce four enhancement strategies--\textit{input image resolution adjustment, data-augmentation, attention mechanisms and an alternative gating function for attention modules}--that can be easily implemented on YOLOv8. We demonstrate that image size enlargement and the proper use of augmentation can lead to enhancement. Additionally, we designed a \textbf{Mixture of Orthogonal Neural-modules Network (MoonNet)} pipeline which consists of multiple attention modules-augmented CNNs. Two well-known attention modules Squeeze-and-Excitation (SE) Block and Convolutional Block Attention Modules (CBAM) were integrated into the backbone of YOLOv8 to form MoonNet design and the MoonNet backbone obtained improved detection accuracy compared to the original YOLOv8 backbone and the single-type attention modules-augmented backbone. MoonNet further proved its adaptability and potential by achieving state-of-the-art performance on the tiny-object benchmark when integrated with the YOLC model. 
Our code is available at:\href{https://github.com/Kihyun11/MoonNet}{https://github.com/Kihyun11/MoonNet}
\end{abstract}
\section{Introduction}
\label{sec:intro}
\vspace{-1em}
\noindent Advances in deep learning have transformed object detection, evolving from early sliding-window CNNs such as R-CNN \cite{girshick2014rcnn} to modern pyramidal, end-to-end architectures. Two-stage detectors—including Faster R-CNN \cite{ren2015faster}, Feature Pyramid Networks (FPN) \cite{lin2017fpn}, and Mask R-CNN \cite{he2017maskrcnn}—first generate region proposals and then refine classification and localization, achieving state-of-the-art accuracy. In parallel, one-stage models prioritize speed by eliminating the proposal stage; seminal examples include SSD \cite{liu2016ssd} and the “You Only Look Once” (YOLO) family \cite{Redmon2016_yolo} \cite{redmon2018yolov3}. Although one-stage detectors excel in training speed, they tend to struggle with detecting tiny objects as a design trade-off. This limitation is especially pronounced in aerial imagery, where targets often occupy only a few pixels amid cluttered backgrounds. Many studies have been conducted to tackle this problem and delivered significant enhancement using YOLO model family. Our research is highly influenced by both Zhu et al\cite{Zhu_2021_ICCV} and Kang et al\cite{Kang2024_SECBAMYOLOv7} which successfully demonstrated the detection accuracy gain in aerial images using the attention modules. Kang et al. especially show that their multiple attention module augmented architecture is better than the SE Block-Only architecture. We further extended this idea to investigate whether the mixed use of attention modules can be always beneficial and what other factors can play an important role in detecting tiny objects using YOLOv8 and YOLC framework.
Our contributions are fourfold: 
(i) effectiveness of higher input resolution, (ii) importance of targeted data augmentation, (iii) compare a hybrid attention-augmented backbone against the single-type attention augmented backbone and obtain the MoonNet design which out-performed among the design candidates, and (iv) suggest a possible alternative gating mechanism for attention modules.

\section{Related Work}
\label{sec:related work}

\noindent \textbf{Tiny object detection in general}: 
Prior work has explored various methods for tiny object detection. Xu et al. \cite{xu2022nwd} suggest that effective enhancement methods for tiny object detection can be categorized into five: \textit{multi-scale feature learning}, \textit{context-based detection}, \textit{data augmentation}, \textit{designing better training strategy} and \textit{label assignment strategy}.
High resolution imagery is crucial as well and numerous approaches were made to improve the resolution of the images such as QueryDet \cite{yang2022querydet}. Some studies such as YOLC \cite{YOLC} even use cropping method to make the training more efficient.

\textbf{Input image resolution}:
Yan et al. \cite{Yan2023} demonstrated that increasing the input image resolution can significantly improve the detection of small-scale objects, as it enables the model to capture more detailed features. Motivated by this, we experiment with multiple input image resolutions in this study to assess their impact. 

\textbf{Data Augmentation}: Proper augmentation is critical when dealing with limited and highly imbalanced datasets. Kisantal et al. \cite{Kisantal2019_dataaug} identified the scarcity of images containing tiny objects and their limited representation in the dataset as a major challenge for small object detection. They addressed this issue using a copy-paste augmentation strategy. 

\textbf{Attention Module \& Gating Function}: Hu et al. \cite{Hu2020} proposed the Squeeze-and-Excitation (SE) block to enhance CNNs by modeling inter-channel dependencies and recalibrating feature responses. Woo et al. \cite{Woo2018} extended this with the Convolutional Block Attention Module (CBAM), which combines channel and spatial attention to better focus on informative regions—an advantage in tiny object detection where spatial precision is crucial. Both modules use sigmoid as a gating function to allow the model to focus on relevant features and avoid vanishing gradients.

\textbf{Previous studies using YOLO}: Plenty of research specifically targeting the enhancement of tiny object detection using the YOLO model already exists. 
Ma et al. \cite{spyolov8s2023} improve YOLOv8s for tiny-object detection by substituting standard strided convolutions with the SPD-Conv module and replacing the PANet neck with SPANet, achieving finer feature preservation and stronger multi-scale fusion.
Hu et al. \cite{elyolo2023} enhance YOLOv5 through an Efficient Spatial Pyramid Pooling (ESPP) module, raising small-object accuracy in aerial imagery while keeping the network lightweight.

\textbf{Relation to prior work}: Implementation of attention modules on the YOLO architecture had been explored previously. Zhu et al. \cite{Zhu_2021_ICCV} introduces TPH-YOLO, which demonstrates the effectiveness of the CBAM with YOLOv5 architecture when detecting objects in aerial imagery. 
Kang et al\cite{Kang2024_SECBAMYOLOv7}'s SE-CBAM-YOLOv7 used both SE Block and CBAM modules on the YOLOv7's architecture to detect tiny aircraft in aerial images.
Unlike Zhu et al. and Kang et al., we focused more on testing several backbone designs with different attention module orientations to check the validity of using multiple attention modules and obtaining the best backbone candidate for enhancement.



We have also made an algorithmic modification on the SE Block and CBAM. As mentioned earlier, sigmoid is a commonly used gating function for the attention modules. A few other modules use sigmoid as well, for instance, ECA-Module \cite{wang2020eca}. In this work, we tested a different gating mechanism $\bigl( 1+\tanh(\cdot)\bigr)$,which is an identity safe gating that can preserve the original feature and also amplifies the scarce relevant features from tiny objects.

\section{Methodology}
\label{sec:methodology}

\subsection{Enhancement Strategy}
\label{enhancement_strategy}
\vspace{-0.5em}
\noindent In total, four different enhancement strategies were implemented on YOLOv8 to obtain efficient training.

\textbf{Varying the input image resolution}: Three different input image resolutions 640, 800, and 928 were tested.

\textbf{Data Augmentation}: A total of three different sets of data augmentation techniques were applied. The first set is the default augmentation setting provided by the Ultralytics. The second set is geometric transformations-focused package and the third set is color space adjustments-focused package.  While these augmentations simulated real-world variations, the primary motivation was to mitigate class imbalance—particularly for under-represented tiny object classes.

\textbf{Attention Modules Integration}: To improve the model’s sensitivity to fine-grained details, attention modules were integrated into the YOLOv8 backbone. Specifically, Squeeze-and-Excitation (SE) blocks and Convolutional Block Attention Modules (CBAM) were inserted directly into select convolutional layers of the backbone. Instead of implementing them as standalone modules, the convolution layer class was extended to perform channel-wise recalibration (SE) and spatial attention (CBAM) inline. 

\textbf{Identity-safe gating mechanism}: While attention modules are implemented on the backbone, the gating function is switched. Thus, one set of two backbones with the same attention module arrangement but different gating was trained and tested. The modifications are presented in Equations\ref{eq:se-conv} and \ref{eq:cbam} and the full mathematical operations of SE Block and CBAM are available via Hu et al.\cite{Hu2020} and Woo et al.\cite{Woo2018}:

\noindent \textbf{Squeeze-Excitation Operation (Ours)}

\begin{subequations}\label{eq:se-conv}
\begin{align}
y_{\text{original}} &= x \odot \sigma(z), \qquad \sigma=\mathrm{sigmoid},                \label{eq:se2:yoriginal}\\[-2pt]
y_{\text{ours}} &= x \odot \bigl(1+\tanh(z)\bigr).                                   \label{eq:se2:yours}
\end{align}
\end{subequations}

\noindent \textbf{CBAM Operation (Ours)}

\begin{subequations}\label{eq:cbam}
\begin{align}
x_c^{\mathrm{original}} &= x \odot \sigma(z_c), \label{eq:cbam:x_original} \\[-2pt]
x_c^{\mathrm{ours}} &= x \odot \bigl(1+\tanh(z_c)\bigr) , \label{eq:cbam:x_ours} \\[2pt]
y_{\mathrm{original}} &= x_c^{\mathrm{original}} \odot \sigma(z_s), \label{eq:cbam:y_original} \\[-2pt]
y_{\mathrm{ours}} &= x_c^{\mathrm{ours}} \odot \bigl(1+\tanh(z_s)\bigr). \label{eq:cbam:y_ours}
\end{align}
\end{subequations}

\textbf{Parameters \& notes}:\;
Channel logits $z$ (SE) and $z_c$ (CBAM) lie in $\mathbb{R}^{C}$ and are broadcast over $H\times W$
(equivalently reshape to $C\times1\times1$). Spatial logits $z_s\in\mathbb{R}^{1\times H\times W}$
are broadcast over channels. Here $\odot$ denotes element-wise multiplication. 
$x$ denotes the input feature map and the subscript $c$ means channel-wise. 
All variables with the \textit{original} subscript denote the original operation whereas that with \textit{ours} denotes the modified operation. 

\textbf{Placement.}\;
For the YOLO-model, the modules were applied to the convolutional layers in the backbone. For the YOLC-model, the attention is applied \emph{per branch at the outputs of the final HRNet stage} (post-branch adapters), not inside
internal convolutional blocks. Let $\{f^{(i)}\}_{i=1}^{B}$ be the multi-resolution maps ($B{=}4$); we compute
$y^{(i)}=\mathcal{A}^{(i)}(f^{(i)})$, $\mathcal{A}^{(i)}\!\in\!\{\text{SE BLOCK},\text{CBAM}\}$. Because gating is
multiplicative, tensor shapes and channels are unchanged, so the YOLC neck/head need no code changes.

\vspace{-1em}
\subsection{Final Configuration}
\label{ssec:final_configuration}
\vspace{-0.5em}
\noindent Following a multi-stage enhancement optimization process, the final model configuration was determined by selecting the most effective combinations of image resolution, data augmentation techniques, network architecture, and gating function. Using the modified backbone with integrated attention modules and the best-performing input image resolution setting, the final model was trained from scratch.
\vspace{-1em}

\section{Experiments and Results}
\label{sec:Experiments and Results}

\subsection{Datasets}
\label{ssec:datasets}
\vspace{-0.5em}
\noindent \textbf{Modified DOTAv2.0.}
We modified the original DOTA 2.0 dataset by selectively retaining only those object categories that consistently exhibit small spatial footprints in aerial imagery. The resulting modified dataset comprises five classes: \textit{small vehicle, large vehicle, ship, plane, and storage tank}. Since the modified dataset was used, the evaluation could not be done through evaluation server but was performed locally using the validation set. The dataset is fully reproducible and the method is provided in the repository. Refer to Table~\ref{tab:dataset_size_comparison}, the retained objects occupy approximately \textbf{2214 pixels² (47×47)}. 

\textbf{VisDrone.} We also evaluated on the VisDrone benchmark in Table~\ref{tab:backbone-attention} and Table~\ref{tab:state-of-the-art} to assess generalization and enable fair comparison. 
Since the VisDrone server is currently unavailable, we used the validation set for evaluation.

\begingroup
\setlength{\textfloatsep}{-5pt}

\begin{figure}[!ht]
    \centering
    \renewcommand{\arraystretch}{1.5}
    \setlength{\tabcolsep}{4pt}
    

    \resizebox{\columnwidth}{!}{
    \begin{tabular}{>{\centering\arraybackslash}p{0.18\textwidth}
                    >{\centering\arraybackslash}p{0.18\textwidth}
                    >{\centering\arraybackslash}p{0.18\textwidth}
                    >{\centering\arraybackslash}p{0.18\textwidth}
                    >{\centering\arraybackslash}p{0.18\textwidth}}
        \large\textbf{Small Vehicle} &
        \large\textbf{Large Vehicle} &
        \large\textbf{Ship} &
        \large\textbf{Plane} &
        \large\textbf{Storage Tank} \\
        
        \includegraphics[width=0.9\linewidth]{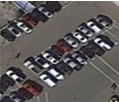} &
        \includegraphics[width=0.9\linewidth]{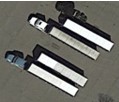} &
        \includegraphics[width=0.9\linewidth]
        {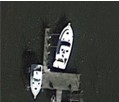} &
        \includegraphics[width=0.9\linewidth]
        {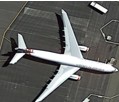} &
        \includegraphics[width=0.9\linewidth]{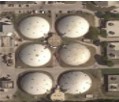} \\
        
        \includegraphics[width=0.9\linewidth]{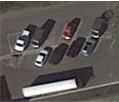} &
        \includegraphics[width=0.9\linewidth]{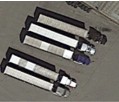} &
        \includegraphics[width=0.9\linewidth]
        {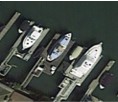} &
        \includegraphics[width=0.9\linewidth]
        {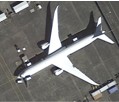} &
        \includegraphics[width=0.9\linewidth]{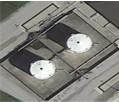  } \\
        
        \includegraphics[width=0.9\linewidth]{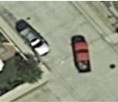} &
        \includegraphics[width=0.9\linewidth]{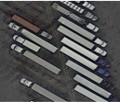} &
        \includegraphics[width=0.9\linewidth]
        {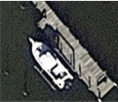} &
        \includegraphics[width=0.9\linewidth]
        {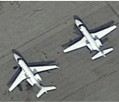} &
        \includegraphics[width=0.9\linewidth]{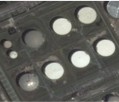  } \\
        
        \includegraphics[width=0.9\linewidth]{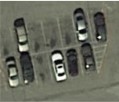} &
        \includegraphics[width=0.9\linewidth]{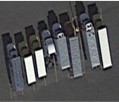} &
        \includegraphics[width=0.9\linewidth]
        {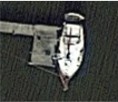} &
        \includegraphics[width=0.9\linewidth]
        {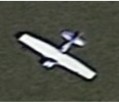} &
        \includegraphics[width=0.9\linewidth]{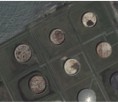  } \\
    \end{tabular}
    }
    \caption{Object categories of the Modified-DOTA}
    \label{modified DOTA}
\end{figure}
 \vspace{-2em}

\begin{table}[!hb]
\centering
\caption{Pixel size comparison}
\label{tab:dataset_size_comparison}
{\scriptsize
\resizebox{\columnwidth}{!}{%
\begin{tabular}{l r}
\toprule
Object Classes & Average pixel size \\
\cmidrule(lr){1-2}
Small Vehicle & \(\approx 1458 \,\text{px}^2 \ (38 \times 38)\) \\ 
Large Vehicle  & \(\approx 12407 \,\text{px}^2 \ (111 \times 111)\) \\
Ship & \(\approx 1199 \,\text{px}^2 \ (35 \times 35)\) \\
Plane & \(\approx 580 \,\text{px}^2 \ (24 \times 24)\) \\
Storage Tank & \(\approx \, 2591 \,\text{px}^2 \ (51 \times 51)\) \\
\cmidrule(lr){1-2}
\multicolumn{2}{l}{Small Object Category Threshold} \\
\cmidrule(lr){1-2}
COCO -``small" \cite{lin2014coco}& \(1024 \,\text{px}^2 \ (32 \times 32)\) \\
\textbf{Modified-DOTA (Ours)} & \(\approx \textbf{2214} \,\textbf{px}^2 \ \textbf{(47} \times \textbf{47)}\) \\
\bottomrule
\end{tabular}
}
}
\end{table}
\endgroup
\vspace{-1em}


\subsection{Training setting}
\label{ssec:setting}
\vspace{-0.5em}

\noindent \textbf{YOLOv8}: All the training runs were conducted on a local workstation equipped with an NVIDIA RTX 4080 SUPER GPU (16 GB VRAM). The environment ran Ultralytics 8.3.239, Python 3.9.10, and PyTorch 2.5.1 on CUDA 12.1

\textbf{YOLC}: All the trainings were conducted in a secure cloud using one NVIDIA A40 GPU. The environment ran PyTorch 2.1.0, Python 3.10, and CUDA 11.8.0 on Ubuntu 22.04.


\textbf{Trial screening}: Input image adjustment, augmentation, and backbone testing were performed over 50 epochs (batch size 4, $640\times640$ image size) using the AdamW optimizer and a modified, default-augmented DOTA dataset. The trainings used a pretrained YOLOv8n-obb model.

\textbf{Final training}: The YOLOv8n-obb with the default and MoonNet backbones were trained for 150 epochs with batch size 4 and the selected candidates setting. The YOLC model trainings were conducted for 50 epochs. 

\textbf{Evaluation metrics}: Performance was evaluated using $AP_{50}$, $AP$, recall, and precision for YOLOv8; and $AP_{50}$, $AP_{75}$, and $AP$ for YOLC in MMDetection. $AP_{50}$ and $AP_{75}$ are obtained at the single IoU threshold 0.5 and 0.75 over all categories. $AP$ is obtained by averaging over threshold 0.5 to 0.95 over all categories.
\vspace{-1em}

\subsection{Input image resolution adjustment}
\label{ssec:hyperparam tuning}
\vspace{-0.5em}
\noindent We identified the image size of (928x928) as yielding the highest performance in $AP_{50}$, $AP$, and Recall.
\vspace{-0.5em}
\begin{table}[h]
\centering
\caption{Performance comparison across image size}
\label{tab:merged-performance-reordered}
\resizebox{\columnwidth}{!}{%
\begin{tabular}{l r r r r}
\toprule
Input image resolution& $AP_{50}$ & $AP$ & Recall & Precision \\
\midrule

(640$\times$640) & 0.621 & 0.436 & 0.539 & 0.796\\
(800$\times$800) & 0.665 & 0.489 & 0.583 & \textbf{0.821}\\
\textbf{(928$\times$928) \cmark} & \textbf{0.696} & \textbf{0.525} & \textbf{0.613} & 0.819\\
\bottomrule
\end{tabular}
}
\end{table}

\vspace{-2em}

\subsection{Data augmentation}
\label{ssec:data augmentation}
\noindent A total of three different augmentation packages were tested and the implemented augmentation types were explained in Table~\ref{tab:augmentation-performance}. \textbf{Default} denotes default augmentation setting provided by the Ultralytics, \textbf{Geo} denotes geometric transformations, and \textbf{Color} denotes HSV color space adjustment. The exact augmentations are provided in the repository.
\vspace{-1em}
\begin{table}[!tbh]
\centering
\setlength{\tabcolsep}{5pt}
\renewcommand{\arraystretch}{1.15}
\caption{Performance metrics across augmentation}
\label{tab:augmentation-performance}
\resizebox{\columnwidth}{!}{%
\begin{tabular}{l ccc rrrr}
\toprule
Package & Default & Geo & Color  & $AP_{50}$ & $AP$ & Recall & Precision \\
\midrule
\textbf{Ver 1} \cmark &  \cmark &  &  & \textbf{0.621} & \textbf{0.436} & \textbf{0.539} & \textbf{0.796} \\
Ver 2  & \cmark & \cmark   &  & 0.604 & 0.416 & 0.530 & 0.763 \\
Ver 3 & \cmark &  & \cmark & 0.616 & 0.428 & 0.536 & 0.786 \\
\bottomrule
\end{tabular}
}
\end{table}
\vspace{-2em}

\subsection{Attention \& gating mechanism}
\label{ssec:attention mechanism}
\begin{figure}[!h]
\centering
\setlength{\tabcolsep}{8pt} 
\renewcommand{\arraystretch}{1.5} 
\resizebox{\columnwidth}{!}{
\begin{tabular}{c c c: c c c c}
\hline

Backbone & 1st design & 2nd design & \multicolumn{1}{c}{3rd design}
& \multicolumn{1}{c}{4th design} & 5th design & 6th design \\
\hline
\raisebox{-0.5cm} 
{\includegraphics[width=0.2\linewidth,height=0.08\linewidth]{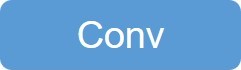}} &\multicolumn{2}{c:}{\raisebox{-0.3em}{\((640 \times 640 \times 3)\)}} &\multicolumn{4}{c}{\raisebox{-0.3em}{\((640 \times 640 \times 3)\)}}     \\ 

& \includegraphics[width=0.2\linewidth,height=0.08\linewidth]{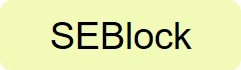} & \includegraphics[width=0.2\linewidth,height=0.08\linewidth]{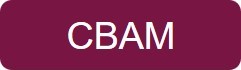}& \includegraphics[width=0.2\linewidth,height=0.08\linewidth]{figures/SEBlock.jpg}& \includegraphics[width=0.2\linewidth,height=0.08\linewidth]{figures/CBAM.jpg} & \includegraphics[width=0.2\linewidth,height=0.08\linewidth]{figures/SEBlock.jpg} & \includegraphics[width=0.2\linewidth,height=0.08\linewidth]{figures/CBAM.jpg}   \\ 

\includegraphics[width=0.2\linewidth,height=0.08\linewidth]{figures/conv_block.jpg} &  \multicolumn{2}{c:}{\raisebox{0.8em}{\((320 \times 320 \times 64 \times w)\)}}&\multicolumn{4}{c}{\raisebox{0.8em}{\((320 \times 320 \times 64 \times w)\)}}    \\ 

& \includegraphics[width=0.2\linewidth,height=0.08\linewidth]{figures/SEBlock.jpg} & \includegraphics[width=0.2\linewidth,height=0.08\linewidth]{figures/CBAM.jpg} &\includegraphics[width=0.2\linewidth,height=0.08\linewidth]{figures/SEBlock.jpg}  &\includegraphics[width=0.2\linewidth,height=0.08\linewidth]{figures/CBAM.jpg}  &\includegraphics[width=0.2\linewidth,height=0.08\linewidth]{figures/CBAM.jpg}  &  
\includegraphics[width=0.2\linewidth,height=0.08\linewidth]{figures/SEBlock.jpg}\\ 

\includegraphics[width=0.2\linewidth,height=0.08\linewidth]{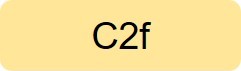} &  &  &  &  &  &  \\ 

\includegraphics[width=0.2\linewidth,height=0.08\linewidth]{figures/conv_block.jpg} &  \multicolumn{2}{c:}{\raisebox{0.8em}{\((160 \times 160 \times 128 \times w)\)}}&\multicolumn{4}{c}{\raisebox{0.8em}{\((160 \times 160 \times 128 \times w)\)}}  \\ 

& \includegraphics[width=0.2\linewidth,height=0.08\linewidth]{figures/SEBlock.jpg} &\includegraphics[width=0.2\linewidth,height=0.08\linewidth]{figures/CBAM.jpg}  & \includegraphics[width=0.2\linewidth,height=0.08\linewidth]{figures/SEBlock.jpg} &\includegraphics[width=0.2\linewidth,height=0.08\linewidth]{figures/CBAM.jpg}  &\includegraphics[width=0.2\linewidth,height=0.08\linewidth]{figures/SEBlock.jpg}  &\includegraphics[width=0.2\linewidth,height=0.08\linewidth]{figures/CBAM.jpg}  \\ 

\includegraphics[width=0.2\linewidth,height=0.08\linewidth]{figures/c2f.jpg} &  &  &  &  &  &  \\ 

\includegraphics[width=0.2\linewidth,height=0.08\linewidth]{figures/conv_block.jpg} &  \multicolumn{2}{c:}{\raisebox{0.8em}{\((80 \times 80 \times 256 \times w)\)}}&\multicolumn{4}{c}{\raisebox{0.8em}{\((80 \times 80 \times 256 \times w)\)}}  \\ 

&  \includegraphics[width=0.2\linewidth,height=0.08\linewidth]{figures/SEBlock.jpg}& \includegraphics[width=0.2\linewidth,height=0.08\linewidth]{figures/CBAM.jpg}& \includegraphics[width=0.2\linewidth,height=0.08\linewidth]{figures/CBAM.jpg} &\includegraphics[width=0.2\linewidth,height=0.08\linewidth]{figures/SEBlock.jpg}  &\includegraphics[width=0.2\linewidth,height=0.08\linewidth]{figures/CBAM.jpg}  &\includegraphics[width=0.2\linewidth,height=0.08\linewidth]{figures/SEBlock.jpg}  \\ 

\includegraphics[width=0.2\linewidth,height=0.08\linewidth]{figures/c2f.jpg} &  &  &  &  &  &  \\ 

\includegraphics[width=0.2\linewidth,height=0.08\linewidth]{figures/conv_block.jpg} &  \multicolumn{2}{c:}{\raisebox{0.8em}{\((40 \times 40 \times 512 \times w)\)}}&\multicolumn{4}{c}{\raisebox{0.8em}{\((40 \times 40 \times 512 \times w)\)}}  \\ 

& \includegraphics[width=0.2\linewidth,height=0.08\linewidth]{figures/SEBlock.jpg}& \includegraphics[width=0.2\linewidth,height=0.08\linewidth]{figures/CBAM.jpg} & \includegraphics[width=0.2\linewidth,height=0.08\linewidth]{figures/CBAM.jpg}& \includegraphics[width=0.2\linewidth,height=0.08\linewidth]{figures/SEBlock.jpg} & \includegraphics[width=0.2\linewidth,height=0.08\linewidth]{figures/SEBlock.jpg} & \includegraphics[width=0.2\linewidth,height=0.08\linewidth]{figures/CBAM.jpg} \\ 

\includegraphics[width=0.2\linewidth,height=0.08\linewidth]{figures/c2f.jpg}& \multicolumn{2}{c:}{\raisebox{0.8em}{\((20 \times 20 \times 1024 \times w)\)}}&\multicolumn{4}{c}{\raisebox{0.3em}{\((20 \times 20 \times 1024 \times w)\)}} \\ 

\includegraphics[width=0.2\linewidth,height=0.08\linewidth]{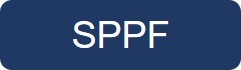}& & & & & & \\ 
\hline
\end{tabular}
 }
\caption{Backbone designs}
\label{tab:backbone designs}
\end{figure}

\noindent A total of six attention-augmented backbones were tested in Table \ref{tab:backbone-attention} and the designs are displayed in Figure \ref{tab:backbone designs}. Dotted lines are used to distinguish between backbones employing single-type versus multiple-type modules. In the trainings using modified DOTA, the third design outperformed other candidates. The top 3 candidates were then trained with VisDrone dataset to check performance and the second design (Only-CBAM) achieved the highest accuracy.

\textbf{The third design}: Although the third design failed to obtain the best performance when trained using VisDrone, we decided to use the third design for the final training configuration since it edges out other backbones when trained with DOTA. We name this particular backbone with SE Block-CBAM arrangement as \textbf{Mixture of Orthogonal Neural-modules Network(MoonNet)}.

Unfortunately, identity-safe gating $(1+\tanh(\cdot)\bigr)$ failed to achieve enhancement when trained with YOLO models. Thus, we did not run additional trainings using tanh gating when using VisDrone. However, tanh gate achieved enhancement when adapted 
to YOLC framework in Table \ref{tab:state-of-the-art} and \ref{tab:gating mechanism comparison}.

\vspace{-1em}
\begin{table}[!hb]
\centering
\caption{Backbone results}
\label{tab:backbone-attention}
\resizebox{\columnwidth}{!}{%
\begin{tabular}{l c c r r r r}
\hline
Backbone & sigmoid & 1+tanh & $AP_{50}$ & $AP$ & Recall & Precision \\
\hline
\multicolumn{7}{l}{Dataset: Modified DOTAv2.0 (50 epochs)}\\
\hline
Default      & \multicolumn{2}{c}{no attention} & 0.486 & 0.299 & 0.427 & 0.690 \\
1st Design   &\cmark & & 0.491 & 0.301 & 0.437 & 0.685\\
             & &\cmark & 0.471 & 0.287 & 0.417  &0.700 \\
2nd Design   &\cmark & & 0.489 & 0.299 & 0.433 & 0.702\\
            & &\cmark & 0.467 & 0.279 & 0.415 & 0.702\\
\textbf{3rd Design} \cmark   &\cmark & & \textbf{0.503} & \textbf{0.313} & \textbf{0.444} & 0.697\\
             & &\cmark & 0.490 &0.300  &0.430  &0.702 \\
4th Design   &\cmark & &0.483  &0.294  &0.441  &0.681 \\
             & &\cmark &  0.473&0.289  &0.427  &0.680 \\
5th Design   &\cmark & & 0.483 & 0.293  &0.425  & \textbf{0.706}   \\
             & &\cmark & 0.472 &0.286  & 0.424 & 0.682\\
6th Design   &\cmark & & 0.485 &0.297  &0.434  &0.685 \\
             & &\cmark & 0.471 &0.282  &0.422  &0.677 \\
\hline
\multicolumn{7}{l}{Dataset: VisDrone (50 epochs)}\\
\hline
1st Design  &\cmark & & 0.294 & 0.170 & 0.303 & \textbf{0.414} \\
\textbf{2nd Design} \cmark   & \cmark & & \textbf{0.296} & \textbf{0.172} & \textbf{0.304} & 0.412\\
3rd Design   & \cmark & & 0.293 & 0.169 & 0.303 & 0.409\\
\bottomrule
\end{tabular}
}
\end{table}

\subsection{Final training and integration with YOLC}
\label{final training}
\vspace{-0.5em}
\noindent The final MoonNet-based model was trained from scratch with optimized input image resolutions and augmentations, showing clear improvements over the default YOLOv8n. In Table~\ref{tab:final-perf}, \textbf{Default} refers to the original backbone, \textbf{Res} to optimized input image resolution, and \textbf{Aug} to optimized augmentations. The number of channels from the model's backbone in the final training is not identical to the numbers presented in Table~\ref{tab:backbone-attention}. This is because all the final trainings in the Table~\ref{tab:final-perf} were conducted using YOLOv8n model scale thus actual number of channels were (16, 32, 64, 128, 256) due to the channel multiplier (0.25). In Table~\ref{tab:gflops}, all parameters from preprocess to end-to-end are measured in milliseconds.
\begin{table}[!tbh]
\centering
\setlength{\tabcolsep}{5pt}
\renewcommand{\arraystretch}{1.15}
\caption{Performance metrics for final models}
\label{tab:final-perf}
\resizebox{\columnwidth}{!}{%
\begin{tabular}{c cccc rrrr}
\toprule
Method & Default & MoonNet & Res & Aug & $AP_{50}$ & $AP$ & Recall & Precision \\
\midrule
Method 1 & \cmark &  &  &  & 0.491 & 0.308 & 0.390 & 0.685 \\
Method 2 & \cmark &  & \cmark &  & 0.596 & 0.411 & 0.501 & 0.763 \\
Method 3 & \cmark &  & \cmark & \cmark & 0.664 &0.485  &0.565  &0.843  \\
\textbf{Method 4} &  & \cmark & \cmark & \cmark & \textbf{0.667} & \textbf{0.486} & \textbf{0.566} & \textbf{0.849} \\
\bottomrule
\end{tabular}
}
\end{table}
\vspace{-1.5em}

\begin{table}[!tbh]
\centering
\setlength{\tabcolsep}{5pt}
\renewcommand{\arraystretch}{1.15}
\caption{GFLOPs and latency comparison}
\label{tab:gflops}
\resizebox{\columnwidth}{!}{%
\begin{tabular}{crrrrr}
\toprule
Method & GFLOPs & Preprocess & Inference & Postprocess & End-to-End  \\
\midrule
Method 1 & 8.3& 0.1 & 1.9 & 2.3 & 4.3 \\
Method 2 & 8.3 &  0.2 & 2.2 & 2.6 &  5.0\\
Method 3 & 8.3 & 0.2 & 2.2 & 2.7 &  5.1 \\
Method 4 & 8.4 & 0.2 & 2.6 & 2.4 & 5.2 \\
\bottomrule
\end{tabular}
}
\end{table}

The MoonNet's (SE Block--CBAM) orientation is then adapted to the YOLC framework to check the MoonNet's adaptability and enhancement potential. MoonNet (sigmoid) and MoonNet (1+tanh) both achieved improved results compare to the original YOLC in Table~\ref{tab:state-of-the-art}. The MoonNet (1+tanh) achieved +6.13\% gain in $AP$ and +3.77\% gain in $AP_{50}$ compare to the MoonNet (sigmoid).

\begin{table}[!tbh]
\centering
\setlength{\tabcolsep}{5pt}
\renewcommand{\arraystretch}{1.15}
\caption{The state-of-the-art comparison on VisDrone}
\label{tab:state-of-the-art}
\resizebox{\columnwidth}{!}{%
\begin{tabular}{l crrr}
\toprule
Method & Backbone  &$AP$ & \textbf{$AP_{50}$}& \textbf{$AP_{75}$} \\
\midrule

ClusDet\cite{ClusDet} & ResNet101 & 0.267 & 0.504 & 0.252 \\
ClusDet\cite{ClusDet} & ResNeXt101 & 0.284 & 0.532 & 0.264 \\

DMNet \cite{DMNet}& ResNet101 & 0.285  & 0.481   & 0.294 \\
DMNet \cite{DMNet}& ResNext101 & 0.294  & 0.493   & 0.306\\

YOLC(k=1) \cite{YOLC} & HRNet & 0.303 & 0.516 & 0.307    \\
\textbf{YOLC(k=1)(Ours)} & \textbf{HRNet + MoonNet(sigmoid)} & \textbf{0.310} & \textbf{0.530} & \textbf{0.311}\\
\textbf{YOLC(k=1)(Ours)} & \textbf{HRNet + MoonNet(1+tanh)} & \textbf{0.329} & \textbf{0.550} & \textbf{0.336}\\
\bottomrule
\end{tabular}
}
\end{table}
\vspace{-1.5em}

\begin{table}[!tbh]
\centering
\setlength{\tabcolsep}{5pt}
\renewcommand{\arraystretch}{1.15}
\caption{Gating mechanism comparison}
\label{tab:gating mechanism comparison}
\resizebox{\columnwidth}{!}{%
\begin{tabular}{l ccrrr}
\toprule
Framework & sigmoid  &1+tanh & \textbf{$AP_{50}$}& \textbf{$AP$} \\
\midrule
\textbf{YOLO-MoonNet (DOTA)} \cmark & \cmark &  & \textbf{0.503} & \textbf{0.313} \\
YOLO-MoonNet (DOTA)&  & \cmark & 0.490 & 0.300 \\
\midrule
YOLC-MoonNet (VisDrone)& \cmark &   &  0.310 &0.530  \\
\textbf{YOLC-MoonNet (VisDrone)} \cmark& &  \cmark &    \textbf{0.329}&\textbf{0.550} \\

\bottomrule
\end{tabular}
}
\end{table}

\begin{figure*}[b] 
    \centering
    \renewcommand{\arraystretch}{1.2}
    \setlength{\tabcolsep}{3pt} 
    

    \resizebox{\textwidth}{!}{
    \begin{tabular}{cccccc}
        \textbf{Method 1} & \textbf{Method 2} & \textbf{Method 3} & \textbf{Method 4} & \textbf{YOLC} & \textbf{YOLC+MoonNet} \\
        
        \includegraphics[width=0.16\textwidth, height=0.2\textwidth]{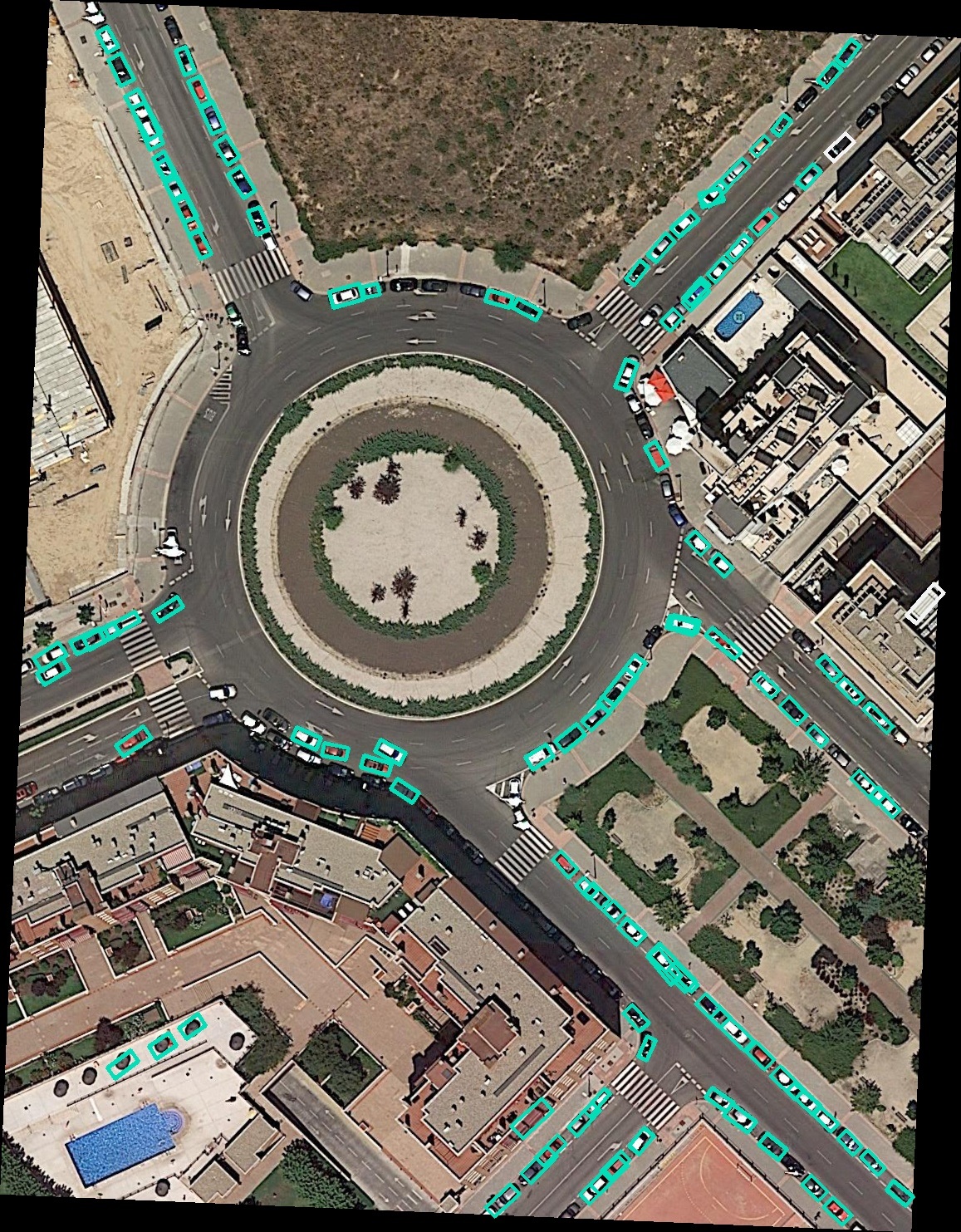} &
        \includegraphics[width=0.16\textwidth, height=0.2\textwidth]{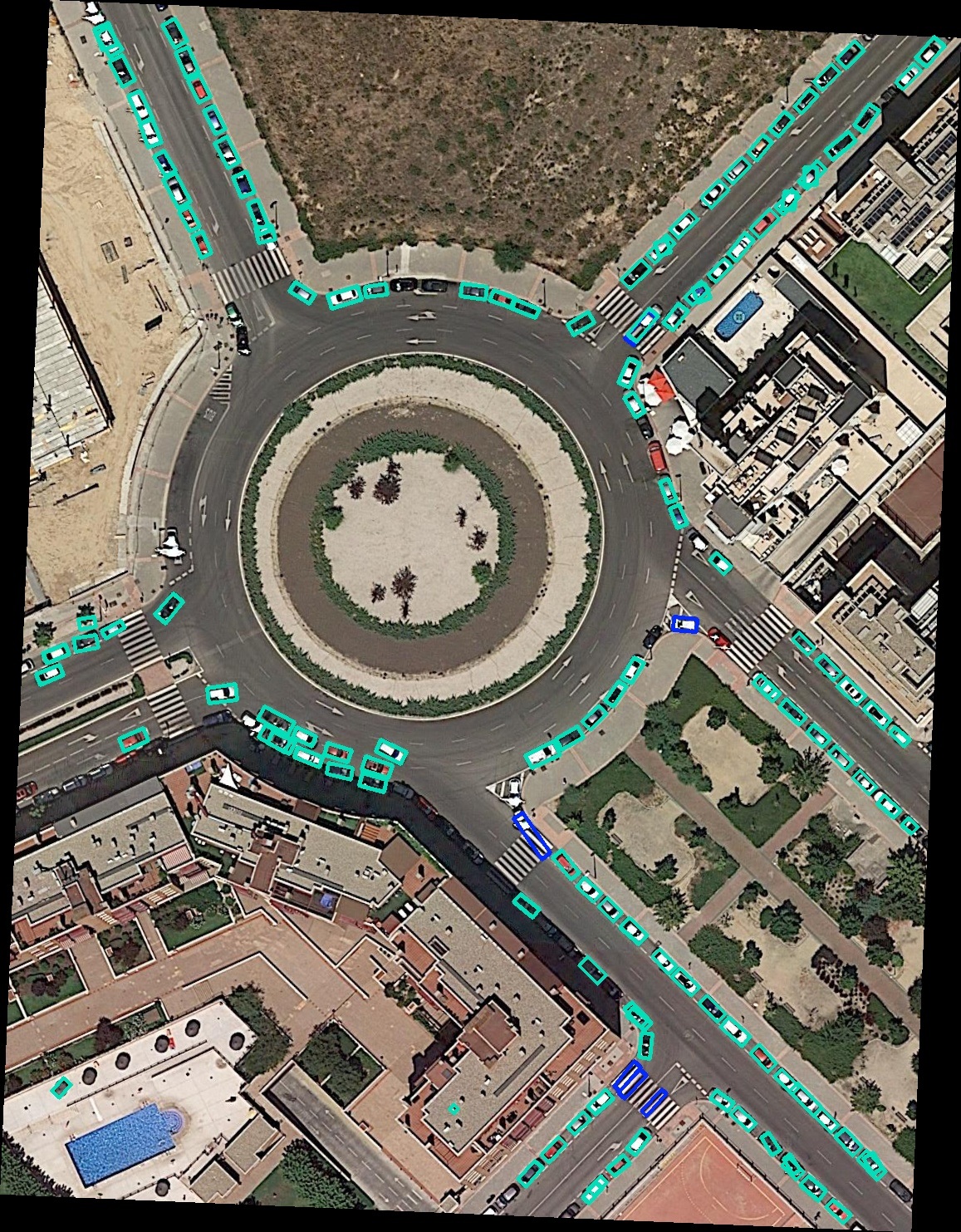} &
        \includegraphics[width=0.16\textwidth, height=0.2\textwidth]{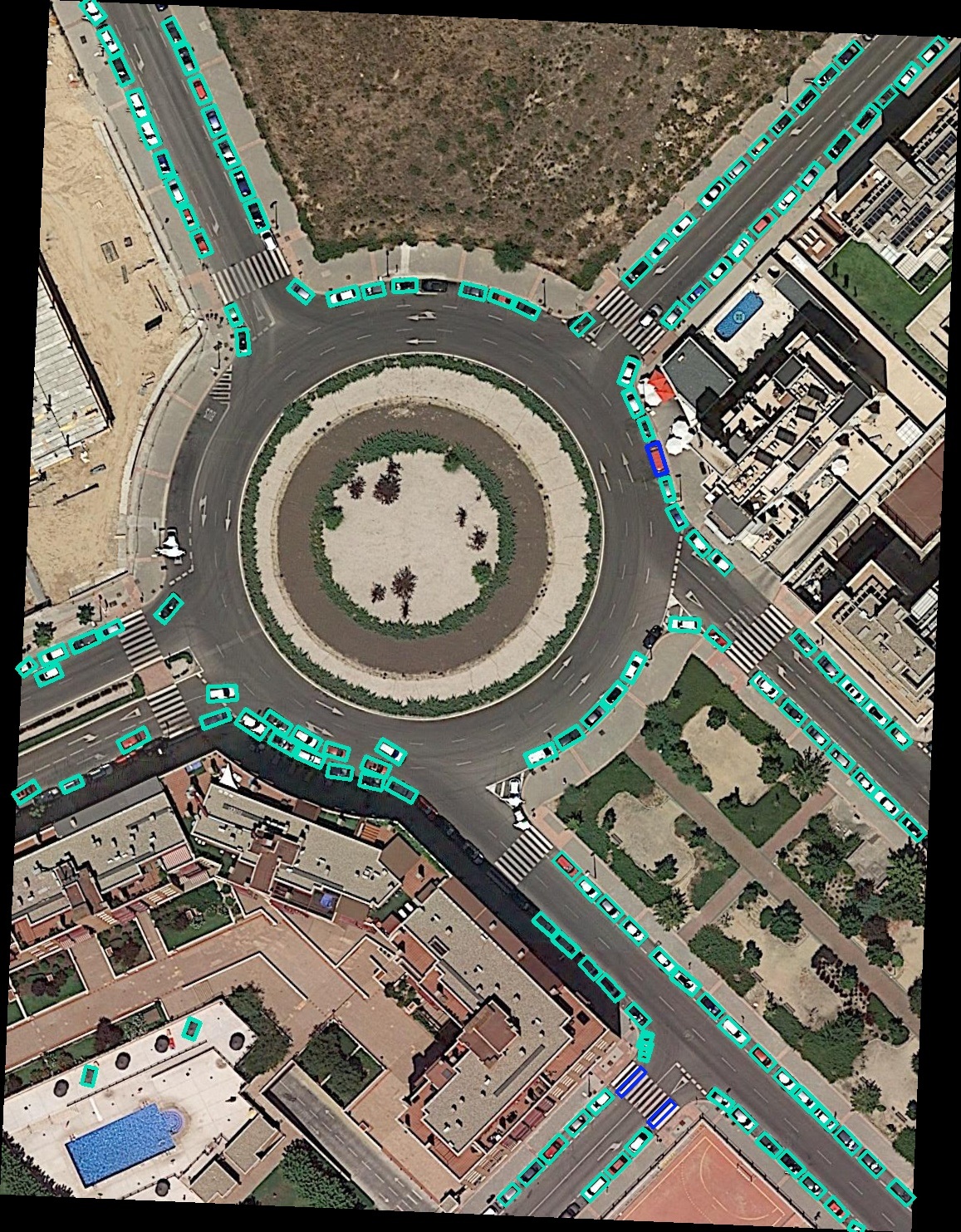} &
        \includegraphics[width=0.16\textwidth, height=0.2\textwidth]{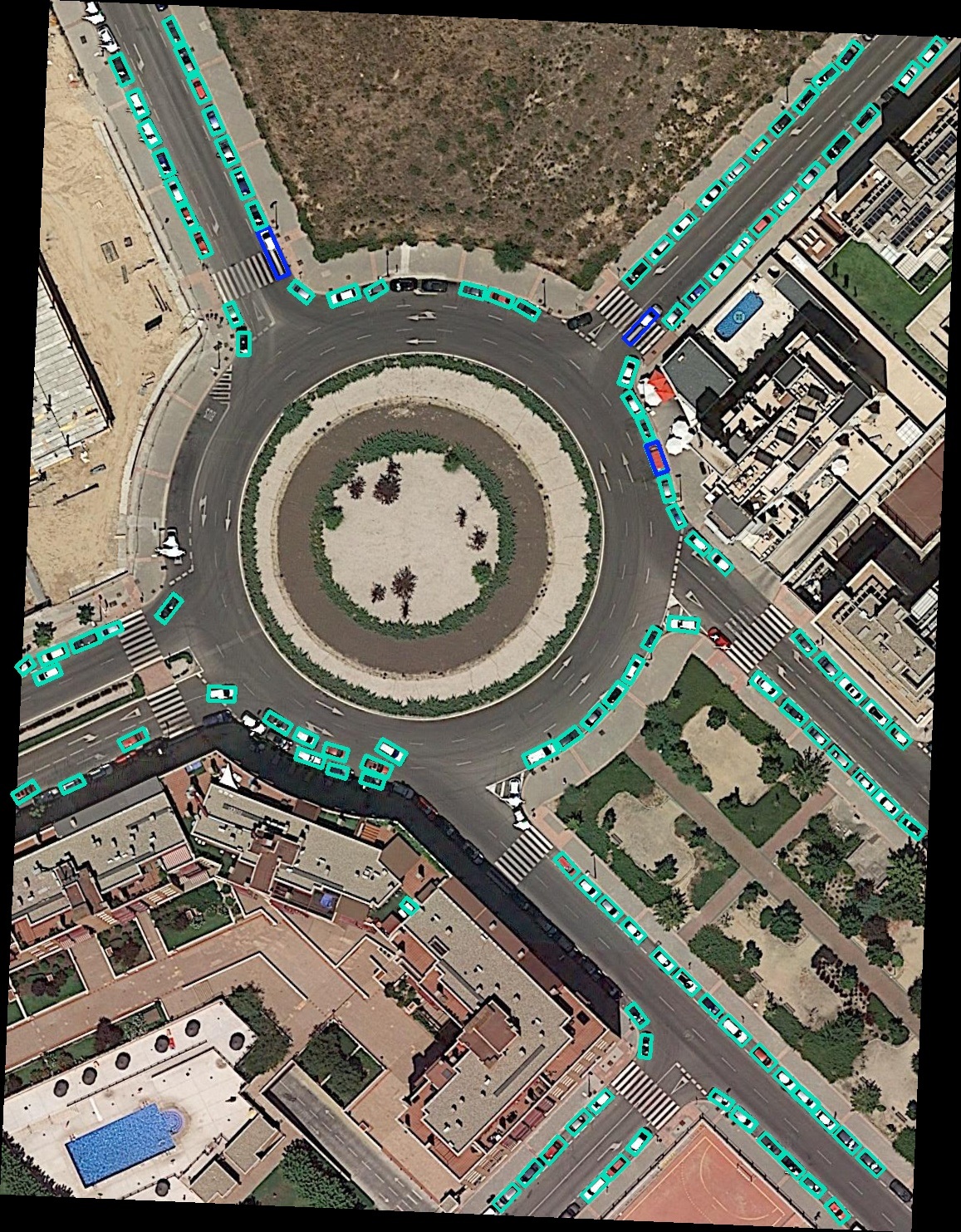} &
        \includegraphics[width=0.16\textwidth, height=0.2\textwidth]{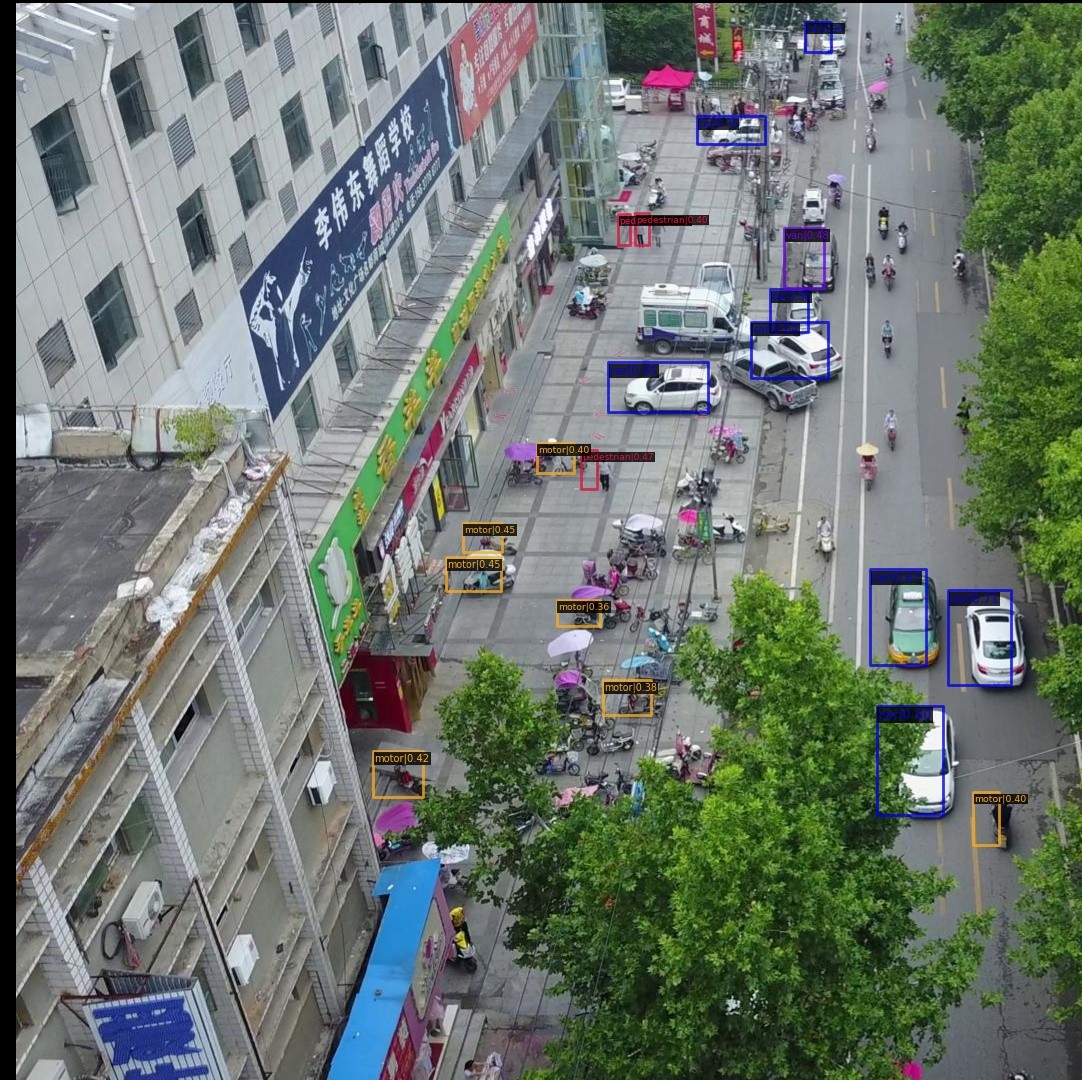} &
        \includegraphics[width=0.16\textwidth, height=0.2\textwidth]{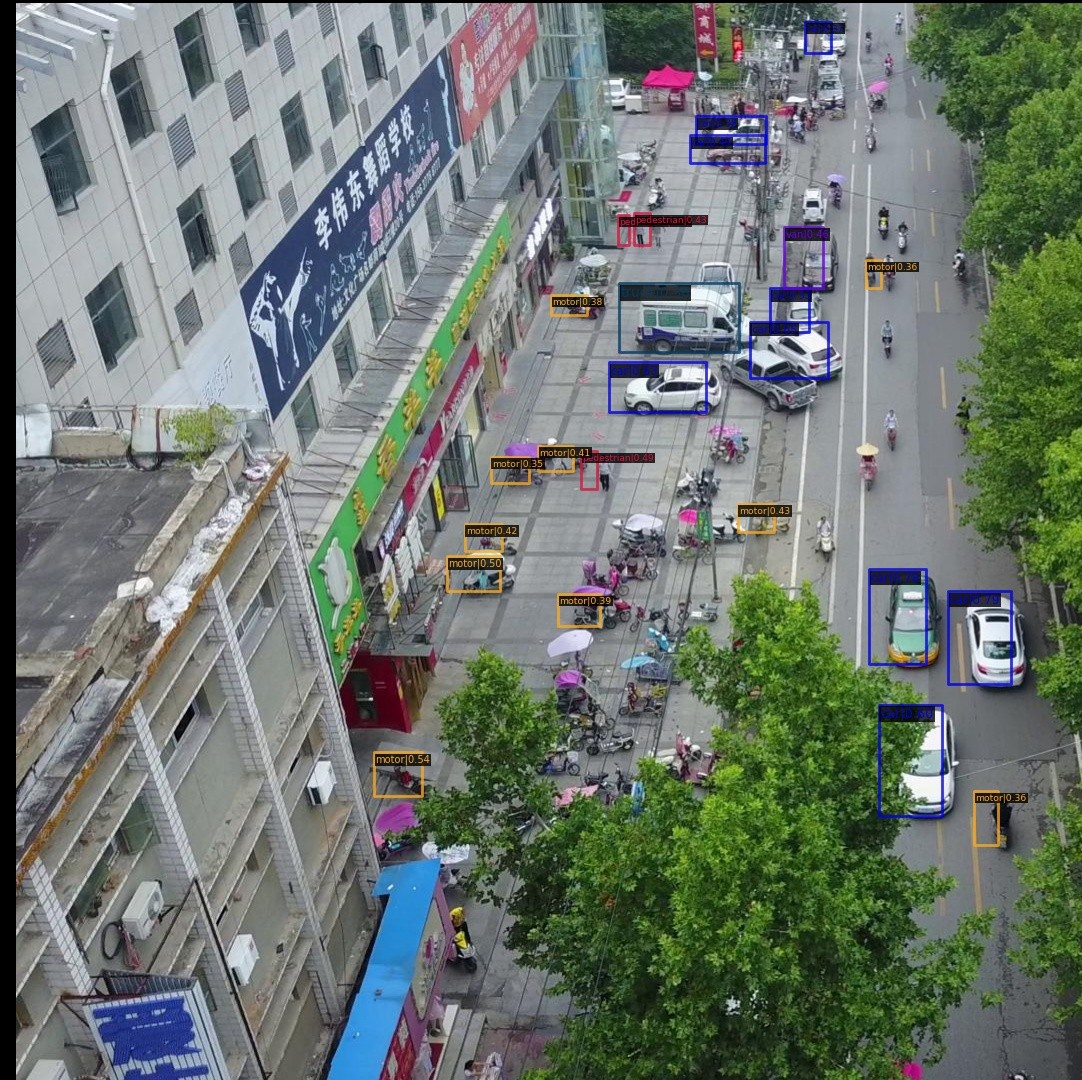} \\
        
        
        \includegraphics[width=0.16\textwidth, height=0.12\textwidth]{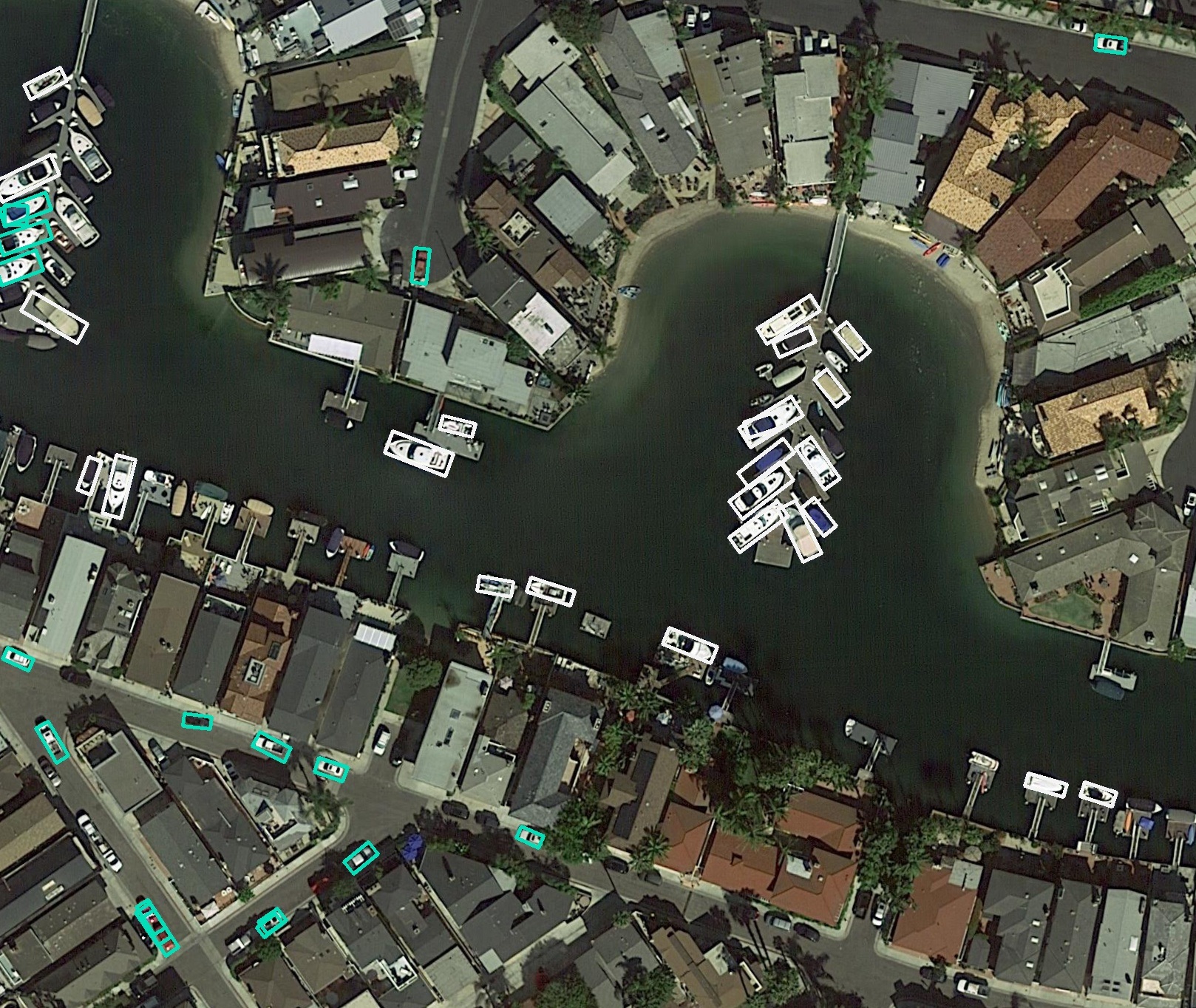} &
        \includegraphics[width=0.16\textwidth, height=0.12\textwidth]{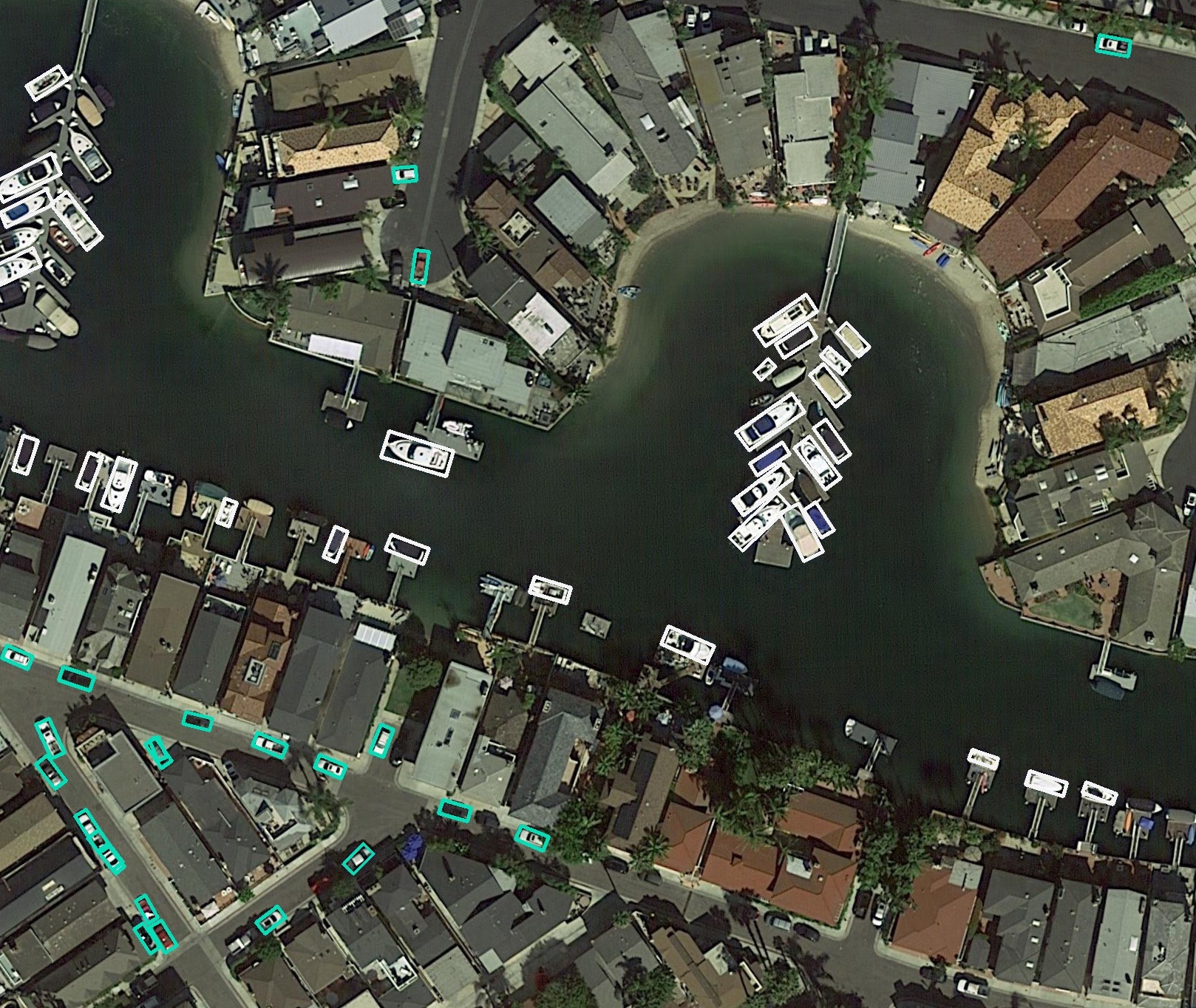} &
        \includegraphics[width=0.16\textwidth, height=0.12\textwidth]{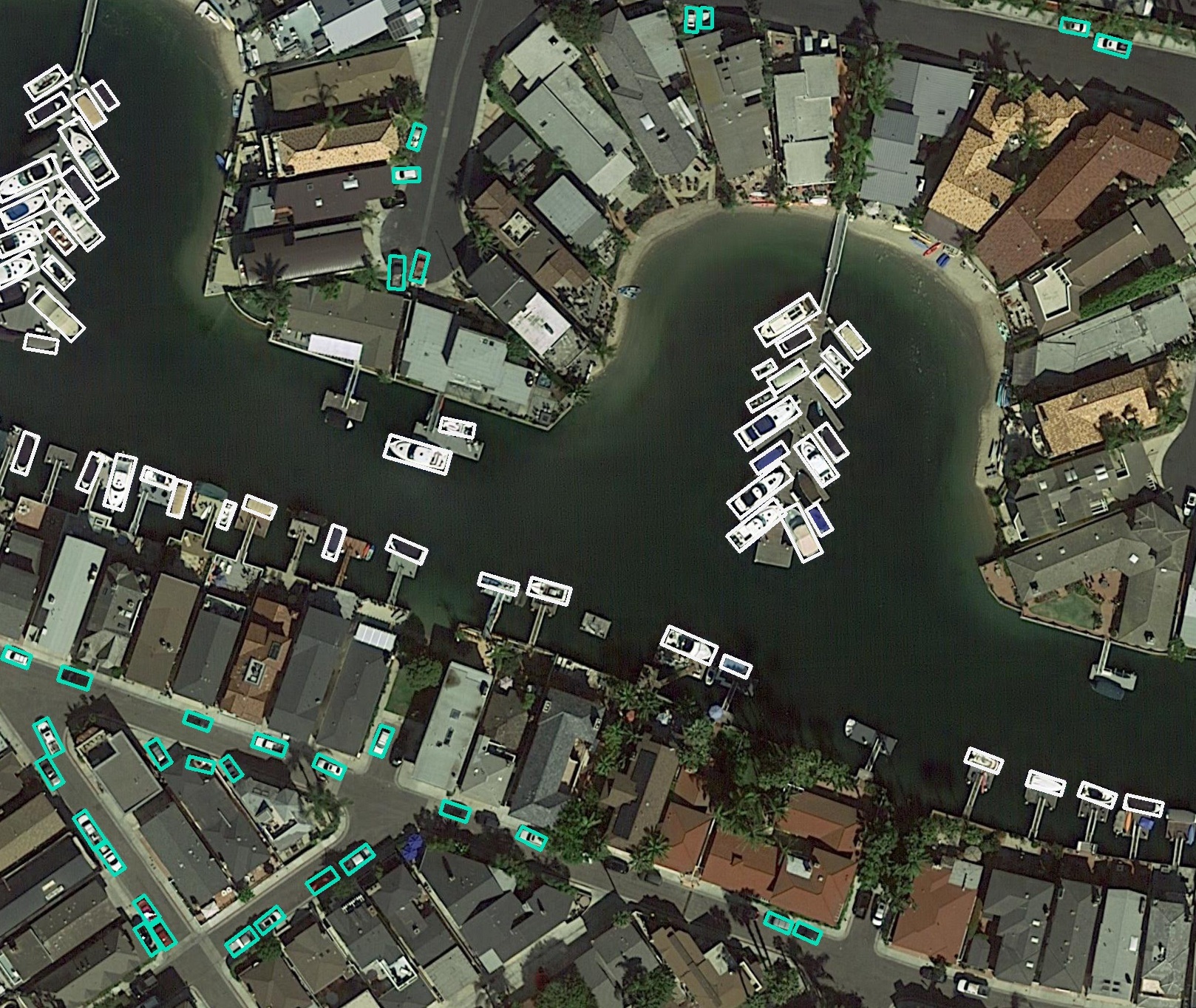} &
        \includegraphics[width=0.16\textwidth, height=0.12\textwidth]{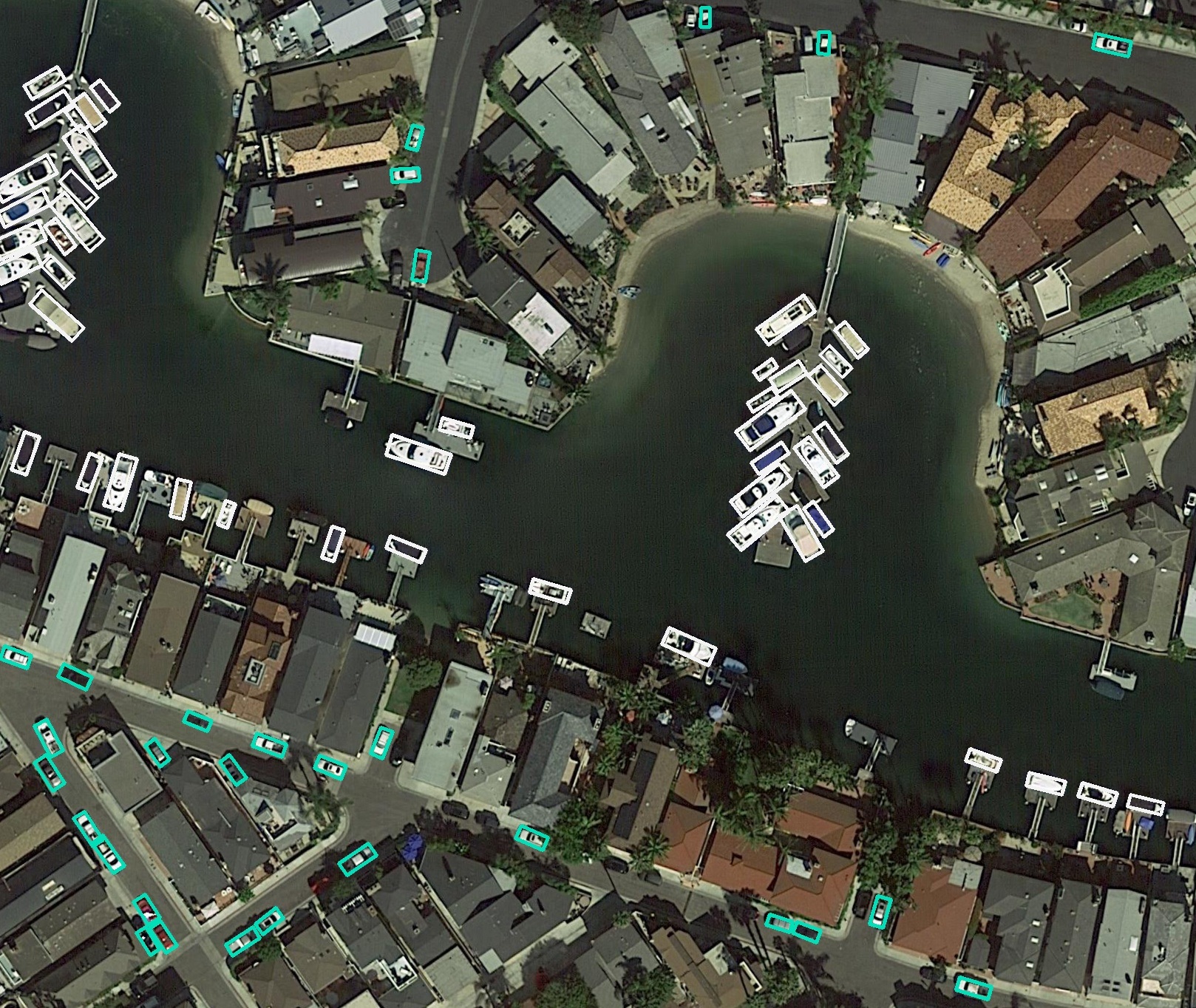} &
        \includegraphics[width=0.16\textwidth, height=0.12\textwidth]{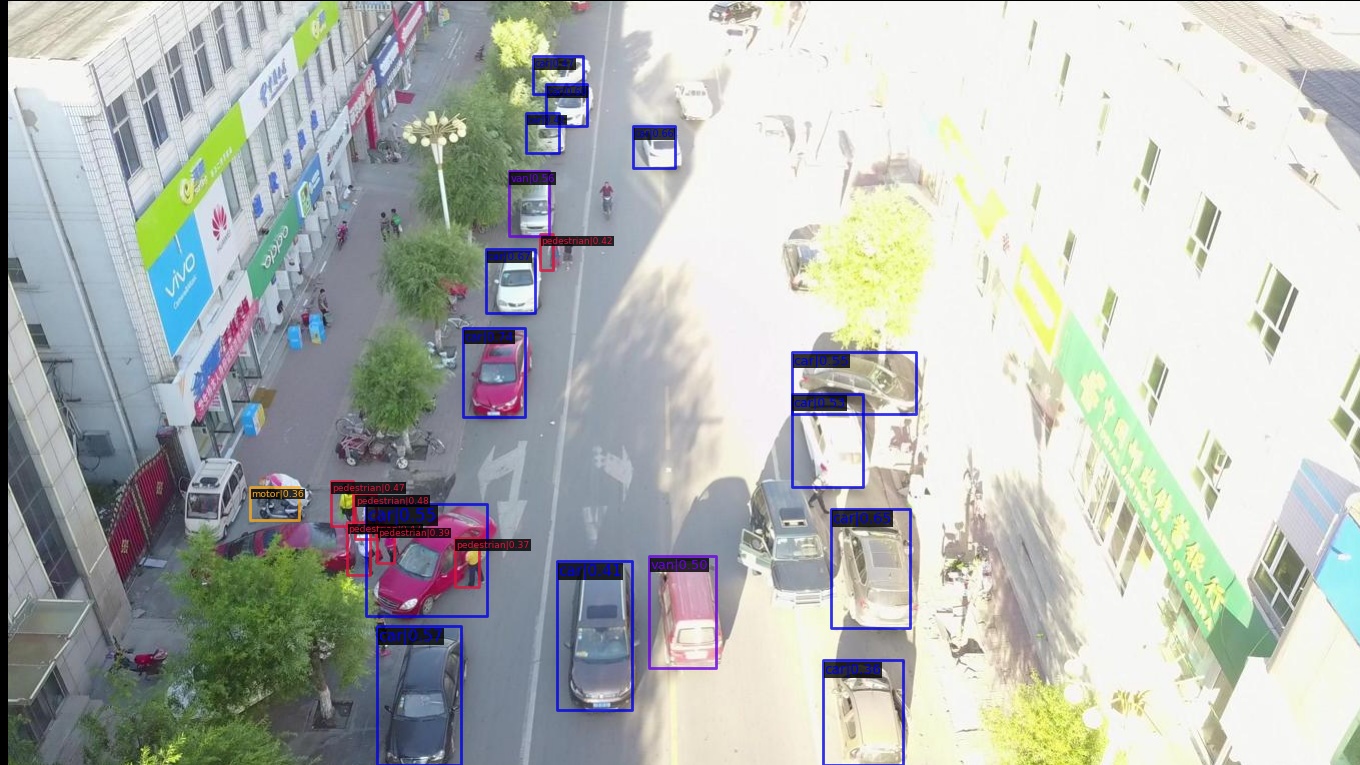} &
        \includegraphics[width=0.16\textwidth, height=0.12\textwidth]{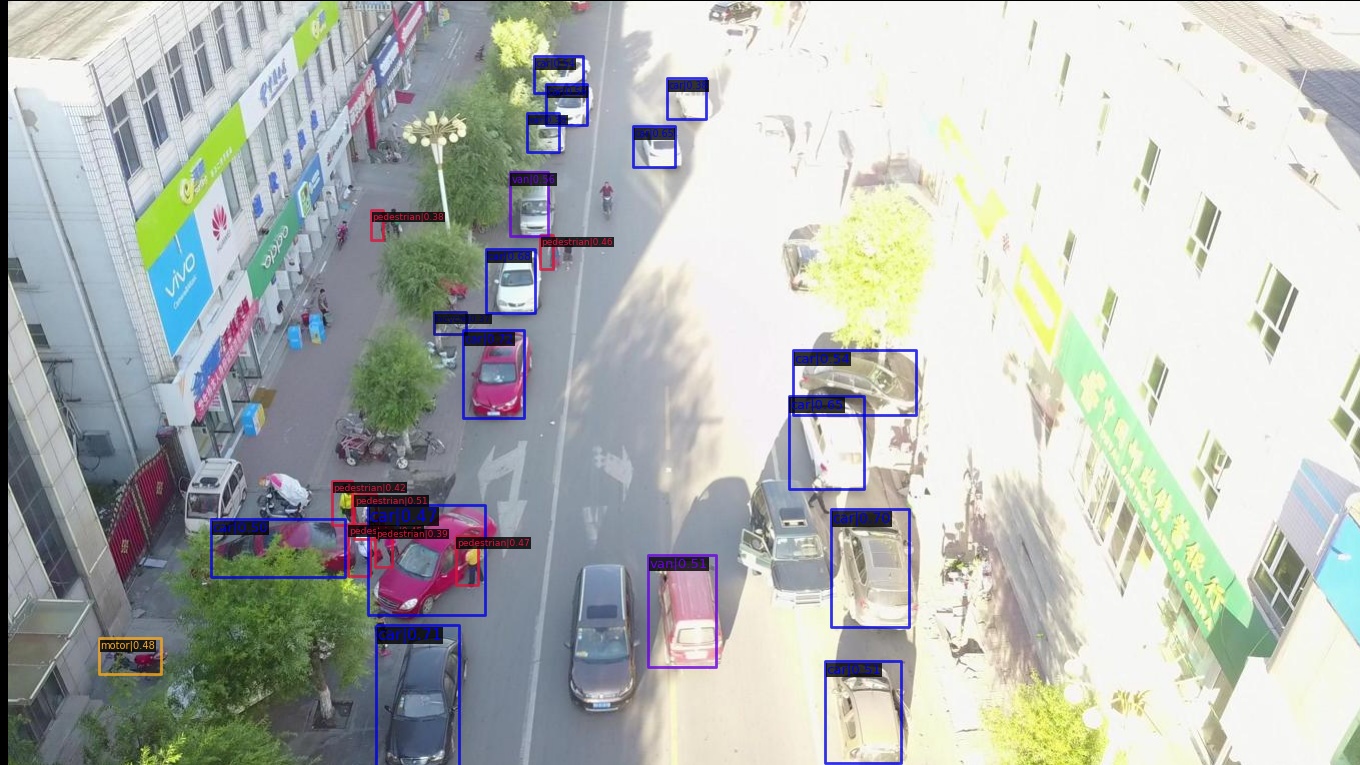} \\
    \end{tabular}
    
    }
    \caption{Visual comparison of detection performance on tiny objects across two different aerial scenes.}
    \label{visual_comparison}
\end{figure*}

\subsection{Ablation Study}
\label{ssec:ablation study}

\noindent\textbf{Image resolution adjustment} \textit{(Method1 $\to$ Method2)} \\
Varying the input image resolution yields +21.38\% gain in $AP_{50}$ and +33.44\% gain in $AP$ showing improved optimization helps recover missed tiny objects.

\textbf{Proper data augmentation} \textit{(Method2 $\to$ Method3)}\\
Instead of adding excessive geometric transformations such as rotation and color space adjustment, mild use of augmentations leads to +11.41\% gain in $AP_{50}$ and +18\% gain in $AP$. 

\textbf{Effectiveness of MoonNet} \textit{(Method3 $\to$ Method4)}\\
Compared to the previous two steps, replacement of the backbone from vanilla CSP to MoonNet yields lower increment. +0.45\% gain in $AP_{50}$ and +0.21\% gain in $AP$, but obtained the highest scores: 0.667 $AP_{50}$ and 0.486 $AP$.

\textbf{Computational burden} \textit{(GFLOPs and latency)}\\
For the training using the final configuration, only 0.1 increase in GFLOPs and 0.9 ms increase were measured from the base score in Table~\ref{tab:gflops} which highlights MoonNet's high deployability with small costs.

\textbf{Hybrid use of attention modules} \textit{(SE Block \& CBAM)} \\
Our third design MoonNet which consists of multiple attention modules demonstrated significant enhancement in detection performance. However, the power of multi-use of attention modules is still questionable. In Table~\ref{tab:backbone-attention}, design 4, 5 and 6 also used SE Block and CBAM together but failed to achieve higher metrics than the designs with single-type attention module. MoonNet's SE Block orientation in the early layer and CBAM in the later stage conveys that the channel emphasis from SE Block followed by spatial focus from CBAM can be particularly effective for combined use of multiple attention modules. When using more than one attention module within a single backbone, careful consideration in module arrangement is required.


\textbf{Adaptability of MoonNet} \textit{(YOLC integration)} \\
In Table~\ref{tab:state-of-the-art}, the MoonNet's attention module arrangement was integrated into the YOLC model with two different gating functions and achieved state-of-the-art performance when compared with ClusDet\cite{ClusDet} and DMNet\cite{DMNet}, both trained on cluster-aware cropped images. 
Ours(sigmoid) obtained +16.10\% gain in $AP$ compared to ClusDet, +8.77\% gain compared to DMNet and +2.31\% gain against the original YOLC.
Ours(1+tanh) obtained +23.22\% gain in $AP$ compared to ClusDet, +15.44\% gain compared to DMNet and +8.58\% gain against the original YOLC.

 \textbf{Alternative gating mechanism} \textit{($1+tanh()$)} \\
 An attention module's gating mechanism is framework-dependent and this is evident through Table~\ref{tab:gating mechanism comparison}'s result. In YOLC framework (HRNet), identity safe gating using tanh might be a better option. Our ablation also revealed that the identity safe gating is unnecessary for the YOLO (CSPDarknet) framework. Therefore, careful consideration on choosing gate function is essential. 

\vspace{-1em}
\subsection{Visual predictions}
\label{ssec:visual predictions}
\vspace{-0.5em}
Visual predictions were made in Figure~\ref{visual_comparison}. As the method evolves from 1 to 4, the model begins to capture more small vehicles, and the method 4 model correctly performs separate detection of small and large vehicles. YOLC+MoonNet model also demonstrates better detection performance.
\vspace{-1em}

\section{Conclusion}
\label{sec:conclusion}
\vspace{-0.5em}
\noindent This research improves tiny object detection by combining resolution adjustment, augmentation, and the MoonNet backbone with tanh gating function. Compared to the base YOLOv8n-obb model, the final model (Method 4) achieved gains of +35.85\% $AP_{50}$, +57.79\% $AP$, +45.13\% recall, and +23.94\% precision.
Moreover, the MoonNet-integrated backbone achieved state-of-the-art performance among the cluster-aware detectors confirming an attention-augmented backbone enhances detection of small targets in aerial imagery. However, the multiple use of attention modules cannot guarantee consistent enhancement. Its orientation and gating function need to be tuned for the appropriate implementation.





\newpage


\bibliographystyle{IEEEbib}
\bibliography{strings,refs}

\end{document}